\documentclass[letterpaper]{article} 
\usepackage{aaai2027}  
\usepackage[hyphens]{url}  
\usepackage{graphicx} 
\usepackage{natbib}  
\usepackage{caption} 
\usepackage{algorithm}
\usepackage{algorithmic}
\usepackage{amsmath}
\usepackage{amssymb}
\usepackage{booktabs}
\usepackage{multirow}
\usepackage{graphicx}
\usepackage[table]{xcolor}
\usepackage{tabularx}
\usepackage{array}
\usepackage{newfloat}
\usepackage{listings}
\DeclareCaptionStyle{ruled}{labelfont=normalfont,labelsep=colon,strut=off} 
\floatstyle{ruled}
\newfloat{listing}{tb}{lst}{}
\floatname{listing}{Listing}

\usepackage{booktabs}

\title{Parameter-Dynamic Adaptive Fusion and Calibration Network for RGBT Tracking}
\author{
    Zhaoding Ding\textsuperscript{\rm 1}, Chenglong Li\textsuperscript{\rm 2}\corresponding, Jiandong Jin\textsuperscript{\rm 3}, Keiwei Ying\textsuperscript{\rm 1}, Wentao Wu\textsuperscript{\rm 1}
}
\affiliations{
    \textsuperscript{\rm 1}School of Artificial Intelligence, Anhui University\\
    \textsuperscript{\rm 2}State Key Laboratory of Opt-Electronic Information Acquisition and Protection Technology, Anhui University\\
    \textsuperscript{\rm 3}School of Computer Science and Technology, Anhui University\\

}

\begin{document}

\maketitle
\begin{abstract}
%

Existing RGBT trackers typically employ fusion functions with fixed parameters across different targets and scenarios. 
Although dynamic-architecture methods improve fusion flexibility by selecting among predefined operations, they still cannot adapt the fusion parameters to the evolving target state.
To address these issues, we propose a Parameter-Dynamic Adaptive Fusion and Calibration Network (PAFCNet) for RGBT tracking.
PAFCNet dynamically generates target-conditioned parameters for multimodal fusion and temporal calibration, enabling the tracking process to adapt to target appearance variations and modality quality fluctuations.
Specifically, we introduce a Target-Adaptive Hypernetwork (TA-HyperNet) that leverages template representations, which preserve stable target identity and recent appearance changes with less background interference, to generate target-conditioned parameters for subsequent fusion and calibration.
Based on TA-HyperNet, we design a target-aware parameter-dynamic fusion module that uses the generated parameters to modulate the fusion process.
This enables the fusion module to adapt to changes in target appearance and complex scene conditions.
Furthermore, since spatio-temporal information propagation may accumulate tracking noise, we propose a dynamic spatio-temporal calibration module that employs TA-HyperNet to generate calibration parameters for spatio-temporal tokens.
By dynamically calibrating historical information before propagation, the module improves the reliability of temporal representations.
Experimental results demonstrate that PAFCNet achieves competitive performance on multiple RGBT tracking benchmarks.
\end{abstract}


\section{Introduction}

\begin{figure}[!ht]
\centering
\includegraphics[width=3.4in]{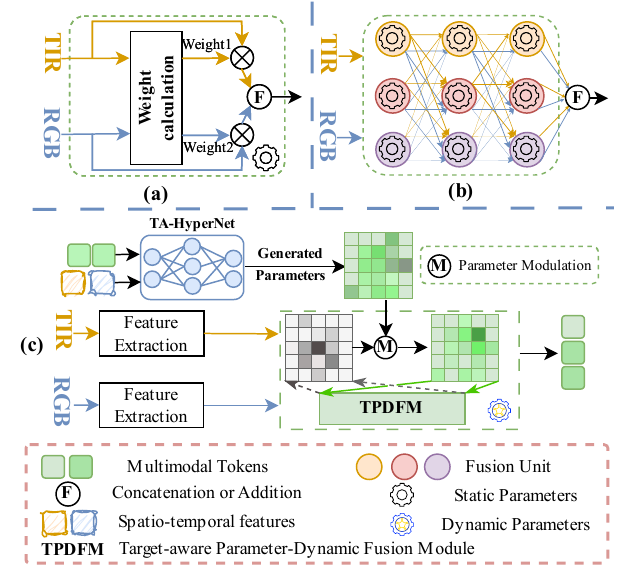}
\caption{(a) Multimodal fusion with fixed parameters and a static architecture. (b) Multimodal fusion with a dynamic architecture. (c) Our proposed parameter-dynamic multimodal fusion method.
}
\label{motivation}
\end{figure}
RGBT tracking exploits the complementary strengths of visible-light (RGB) and thermal-infrared (TIR) modalities to achieve robust all-day target tracking~\cite{ding2025quality,tbsi,jin2026progressive}.
Existing methods typically design fusion architectures with fixed parameters, applying the same fusion function across different targets and tracking scenarios.
They have limited ability to adapt to variations in target appearance and environmental conditions.
Although recent dynamic-architecture methods improve fusion flexibility by selecting or combining predefined fusion operations, their adaptability remains constrained by a fixed candidate space, and the fusion parameters cannot be directly adjusted according to the evolving target state.
Therefore, dynamically adapting the parameters of the multimodal fusion function to  the evolving target state remains insufficiently explored in RGBT tracking.

Existing RGBT tracking methods can be broadly categorized into cross-modal interaction, quality-aware fusion, and dynamic-architecture fusion. 
Cross-modal interaction methods~\cite{tbsi,BAT2024,ainet} integrate information from RGB and TIR modalities through attention mechanisms, feature enhancement, or multi-level information propagation.
For instance, AINet~\cite{ainet} introduces a linear-complexity differential Mamba module for multimodal fusion and hierarchical interaction. Quality-aware methods~\cite{QAT2023,ding2025quality} estimate modality reliability and adaptively regulate modality contributions. For example, QAT~\cite{QAT2023} employs a quality assessment network to predict reliability weights under pseudo-label supervision, as shown in Fig.~\ref{motivation}(a).
Dynamic-architecture methods~\cite{AFTER,li2024dynamic} instead use routing mechanisms to select or combine fusion paths from predefined candidate operations. AFTER~\cite{AFTER} adaptively constructs fusion paths according to input features, as shown in Fig.~\ref{motivation}(b).
Despite these advances, existing methods either share fixed fusion parameters across targets and scenarios or search within a limited predefined operation space, making it difficult to adapt the fusion strategy to dynamic scene and target appearance variations.
In addition, existing tracking methods~\cite{TATrack,ding2025quality,sttrack} have begun to exploit spatio-temporal information to enhance target representation and tracking performance. For example, QSTNet~\cite{ding2025quality} introduces a multimodal spatio-temporal token propagation mechanism to continuously propagate multimodal spatio-temporal information.
However, spatio-temporal information is inevitably affected by noise during propagation, but this issue remains overlooked by existing methods.
To address the above issues, we propose a novel Parameter-Dynamic Adaptive Fusion and Calibration Network (PAFCNet), which introduces a target-adaptive hypernetwork that dynamically generates multimodal fusion parameters conditioned on template representations, enabling the fusion process to adapt to the target state, as shown in Fig.~\ref{motivation}(c).
PAFCNet consists of three core components: a Target-adaptive Hypernetwork (TA-HyperNet), a Target-aware Parameter-Dynamic Fusion Module (TPDFM), and a Dynamic Spatio-temporal Calibration Module (DSCM).
Specifically, TA-HyperNet is designed to dynamically generate target-related parameters for multimodal fusion and spatio-temporal calibration in RGBT tracking.
Within TPDFM, multimodal template features are fed into TA-HyperNet as target-aware cues to generate fusion-specific parameters. These parameters dynamically modulate the multimodal fusion process, allowing it to accommodate target appearance variations and mitigate interference in complex dynamic scenes.
Meanwhile, DSCM is proposed to ensure the reliability of spatio-temporal information during propagation, which generates calibration parameters from template representations and calibrates the spatio-temporal features before propagation, thereby suppressing noise interference and preserving accurate target information.
Our contributions are summarized as follows:
\begin{itemize}
\item We propose a novel parameter-dynamic adaptive fusion and calibration network for RGBT tracking, which employs a target-adaptive hypernetwork to generate multimodal fusion parameters, enabling the fusion process to adapt to target appearance variations and interference in complex dynamic scenes.
\item We design a target-aware parameter-dynamic multimodal fusion module that employs TA-HyperNet to generate the parameters of the fusion module, enabling the fusion process to adapt to the target state and predict reliable modality weights.
\item We design a dynamic spatio-temporal calibration module that employs TA-HyperNet to dynamically generate calibration parameters, enabling the spatio-temporal tokens to be adaptively calibrated before propagation, thereby suppressing noise interference and preserving reliable target information.
\item Extensive experiments on multiple RGBT tracking benchmarks demonstrate that our method achieves competitive performance.
\end{itemize}

\section{Related Work}
\subsection{RGBT Tracking Methods}
Existing RGBT tracking methods can be broadly categorized into cross-modal interaction, quality-aware fusion, and dynamic-architecture fusion~\cite{AFTER,xiang2025cross,moetrack}.
Cross-modal interaction methods~\cite{catpp,zhang2021learning,li2024dynamic} exploit RGB--TIR complementarity through feature disentanglement, enhancement, or multi-level information exchange.
For example, CAT++~\cite{catpp} decomposes target representation learning into multiple challenge-aware branches.
Quality-aware methods~\cite{QAT2023,ding2025quality,TUMFNet} estimate modality reliability and adaptively regulate modality contributions.
TUMFNet~\cite{TUMFNet}, for instance, models modality uncertainty and evaluates dynamic-template reliability.
Dynamic-architecture methods~\cite{AFTER,li2024dynamic,moetrack} improve fusion flexibility by selecting fusion paths, fusion units, or experts according to the input.
AFTER~\cite{AFTER} constructs a candidate fusion space and dynamically organizes fusion units through a router.

Despite these advances, existing methods either retain fixed fusion parameters or select from predefined operations, limiting their ability to adapt the fusion function to target-specific state variations.
In contrast, our method uses template features as target-state cues to dynamically generate fusion parameters, enabling target-conditioned multimodal fusion.

\begin{figure*}[!ht]
\centering
\includegraphics[width=6.5in]{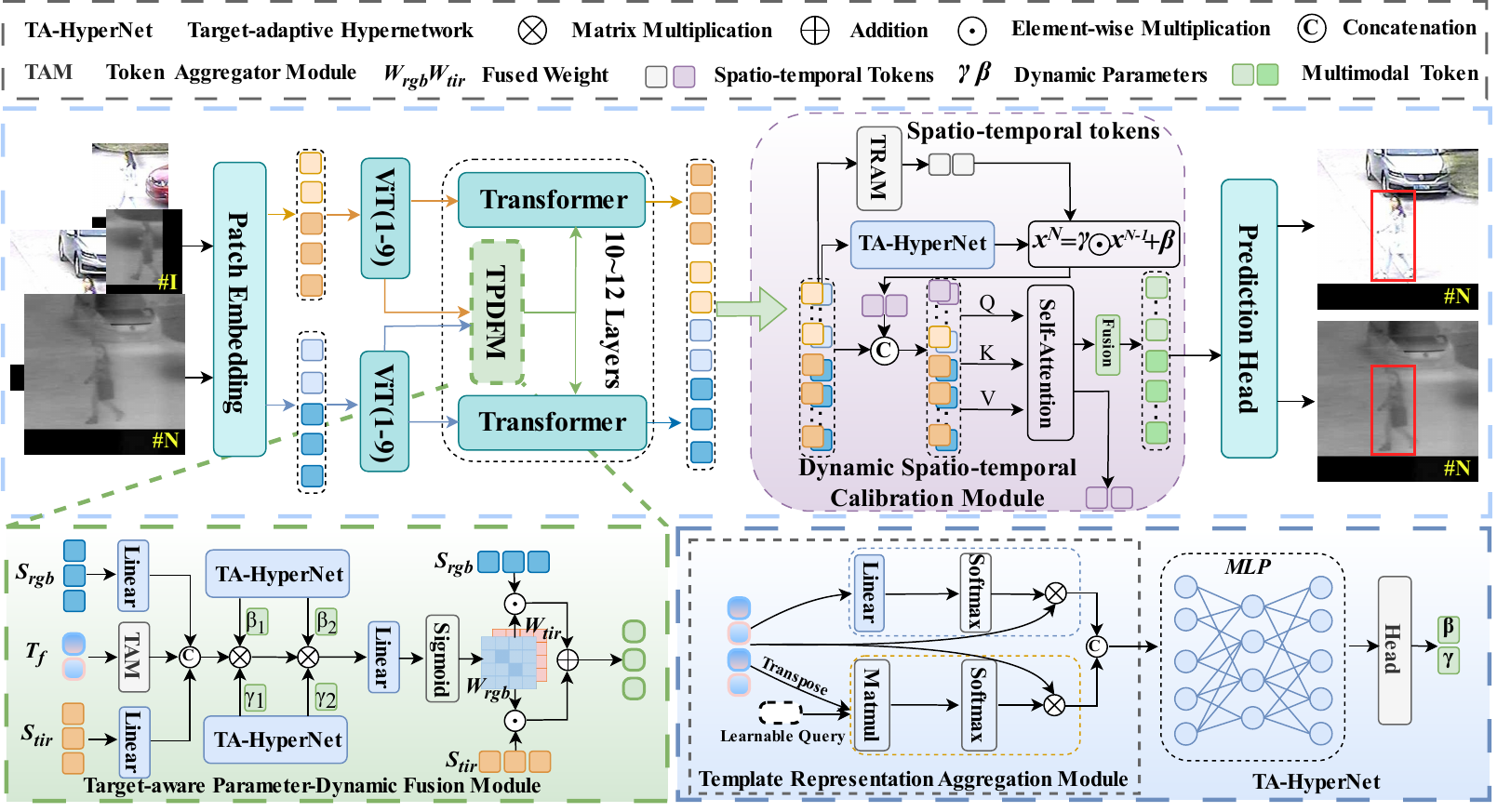}
\caption{The overall framework of our method. TRAM denotes the template representation aggregation module. 
}
\label{network}
\end{figure*}

\subsection{Hypernetworks}
HyperNetworks are typically used to generate all or part of the parameters of a model based on contextual information, thereby enhancing its performance on the current task.
In recent years, hypernetworks have attracted widespread attention due to their efficiency and flexibility, with numerous works across different domains leveraging them to improve model performance~\cite{gwilliam2026design,xu2025sportsal,lv2024hyperlora}.
For example, in natural language processing, HyperLoRA~\cite{lv2024hyperlora} is proposed to generate LoRA parameters from task-oriented information, thereby improving cross-task generalization. Text-to-LoRA~\cite{charakorn2025text} is further introduced, where natural-language task descriptions are mapped to task-specific LoRA adapters in a single forward pass. 
In the medical domain, HyperFusion~\cite{duenias2025hyperfusion} employs a hypernetwork to generate parameters conditioned on tabular clinical information, thereby adapting image processing for multimodal integration.

Although existing hypernetwork-based methods have demonstrated strong flexibility in parameter generation and task adaptation, their potential for addressing the unique challenges of RGBT tracking remains underexplored.
Unlike other works, RGBT tracking requires adaptive cross-modal fusion and reliable spatio-temporal modeling.
To this end, we design a target-adaptive hypernetwork to generate the key parameters required for multimodal fusion and spatio-temporal information calibration.

\section{Methodology}
\subsection{Overview}
The overall architecture of the proposed PAFCNet is illustrated in Fig.~\ref{network}. PAFCNet comprises three core components: the Target-adaptive Hypernetwork (TA-HyperNet), the Target-aware Parameter-Dynamic Fusion Module (TPDFM), and the Dynamic Spatio-temporal Calibration Module (DSCM). 
TA-HyperNet provides a unified target-conditioned parameter-generation
mechanism for TPDFM and DSCM, supporting adaptive multimodal fusion and reliable spatio-temporal information propagation, respectively.
Building on TA-HyperNet, we propose a simple yet effective TPDFM to adapt the multimodal fusion process to evolving target states. TPDFM employs TA-HyperNet to generate target-conditioned fusion parameters, which dynamically modulate multimodal feature integration. TPDFM is inserted into the 10th, 11th, and 12th Transformer layers of the ViT backbone.
Furthermore, to ensure the reliability of spatio-temporal tokens, we develop DSCM based on TA-HyperNet, which calibrates spatio-temporal information during tracking to suppress noise interference and preserve accurate target representations.
%

%

\subsection{Target-adaptive Hypernetwork}
\label{sec:ta_hypernet}

Existing multimodal fusion methods commonly employ shared parameters across different targets and tracking scenarios~\cite{AFTER,ding2025quality}.
However, target appearance and modality reliability may vary substantially throughout tracking, making fixed fusion parameters insufficient for modeling target-specific variations.
To address this limitation, we introduce a Target-adaptive Hypernetwork (TA-HyperNet), as illustrated in Fig.~\ref{network}.
TA-HyperNet extracts complementary global and local target representations from template tokens and generates target-conditioned modulation parameters for multimodal fusion and spatio-temporal calibration.

Given the initial template tokens
$\mathbf{T}^{i}\in\mathbb{R}^{N_i\times C}$
and dynamic template tokens
$\mathbf{T}^{d}\in\mathbb{R}^{N_d\times C}$,
we construct and normalize the template representation as
\begin{equation}
\mathbf{T}
=
[\mathbf{T}^{i};\mathbf{T}^{d}]
\in\mathbb{R}^{N_t\times C},
\qquad
\overline{\mathbf{T}}
=
\operatorname{LN}(\mathbf{T}),
\end{equation}
where $N_t=N_i+N_d$ and $C$ denotes the feature dimension.
For clarity, the batch dimension is omitted.

Directly averaging all template tokens may introduce background noise and weaken discriminative target cues.
We therefore design a Template Representation Aggregation Module (TRAM), consisting of a global content-aware aggregation branch and a multi-query local aggregation branch.

In the global branch, a learnable projection estimates token importance and aggregates the normalized template tokens:
\begin{equation}
\boldsymbol{\alpha}
=
\operatorname{Softmax}
(\overline{\mathbf{T}}\mathbf{W}_{g}),
\qquad
\mathbf{z}_{g}
=
\boldsymbol{\alpha}^{\mathrm{T}}\overline{\mathbf{T}}
\in\mathbb{R}^{C},
\end{equation}
where
$\mathbf{W}_{g}\in\mathbb{R}^{C\times 1}$
is a learnable projection matrix and
$\boldsymbol{\alpha}\in\mathbb{R}^{N_t}$
denotes the normalized token-level importance weights.
This branch summarizes the overall target appearance while emphasizing informative template tokens.

To preserve diverse local target cues, the local branch introduces
$K$ learnable summary queries and performs query-guided aggregation:
\begin{equation}
\begin{aligned}
\mathbf{Q}
&\in\mathbb{R}^{K\times C},
\qquad
\mathbf{A}
=
\operatorname{Softmax}
\left(
\frac{\mathbf{Q}\overline{\mathbf{T}}^{\mathrm{T}}}{\sqrt{C}}
\right),\\
\mathbf{Z}_{l}
&=
\mathbf{A}\overline{\mathbf{T}}
\in\mathbb{R}^{K\times C}.
\end{aligned}
\end{equation}
where
$\mathbf{A}\in\mathbb{R}^{K\times N_t}$
denotes the query-to-template attention weights.
Different queries attend to complementary target patterns, allowing TRAM to preserve multiple local appearance summaries.

The global and local representations are concatenated and mapped into a compact latent representation:
\begin{equation}
\begin{aligned}
\mathbf{u}
&=
[\mathbf{z}_{g};\mathbf{Z}_{l}]
\in\mathbb{R}^{(K+1)C},\\
\mathbf{h}
&=
\phi_s\left(\phi_f(\mathbf{u})\right),
\end{aligned}
\end{equation}
where $\phi_f(\cdot)$ and $\phi_s(\cdot)$ denote linear projections.

Finally, parameter-generation heads produce two groups of channel-wise modulation parameters:
\begin{equation}
\begin{aligned}
\boldsymbol{\gamma}_{j}
&=
\tanh
\left(
\phi_{\gamma_j}(\mathbf{h})
\right),\\
\boldsymbol{\beta}_{j}
&=
\tanh
\left(
\phi_{\beta_j}(\mathbf{h})
\right),
\qquad
j\in\{1,2\}.
\end{aligned}
\end{equation}
where
$\boldsymbol{\gamma}_{j},\boldsymbol{\beta}_{j}\in\mathbb{R}^{H}$,
$H$ denotes the hidden feature dimension.
This bounded parameterization prevents excessive target-conditioned perturbations and stabilizes optimization.

In our implementation, we set $K=4$ and $H=256$.
Thus, TA-HyperNet transforms the global and local appearance cues encoded in the initial and dynamic templates into target-conditioned parameters, enabling subsequent modules to adapt their feature modeling to the current target state.

\subsection{Target-aware Parameter-Dynamic Fusion Module}
\label{sec:TPDFM}

Building upon TA-HyperNet, we develop a Target-Aware Parameter-Dynamic Fusion Module (TPDFM), as illustrated in Fig.~\ref{network}.
TPDFM conditions the multimodal fusion process on the template representation and uses the dynamically generated parameters to modulate the hidden features of the fusion network.
The resulting target-conditioned features are subsequently used to predict token-wise RGB and TIR weights.
Therefore, although the basic fusion network is shared across different targets, its effective feature transformation is dynamically adapted to the current target state.

Let
$\mathbf{S}_{\mathrm{rgb}},\mathbf{S}_{\mathrm{tir}}\in\mathbb{R}^{N_s\times C}$
denote the RGB and TIR search features, respectively, and let
$\mathbf{T}_{f}\in\mathbb{R}^{N_t\times C}$
denote the multimodal template tokens.
We first project the two search features into a low-dimensional interaction space:
\begin{equation}
\widetilde{\mathbf{S}}_{\mathrm{rgb}}
=
\phi_{\mathrm{rgb}}(\mathbf{S}_{\mathrm{rgb}})
\in
\mathbb{R}^{N_s\times C_p},
\end{equation}
\begin{equation}
\widetilde{\mathbf{S}}_{\mathrm{tir}}
=
\phi_{\mathrm{tir}}(\mathbf{S}_{\mathrm{tir}})
\in
\mathbb{R}^{N_s\times C_p},
\end{equation}
where $\phi_{\mathrm{rgb}}(\cdot)$ and $\phi_{\mathrm{tir}}(\cdot)$ are modality-specific linear projections and $C_p$ denotes the projected feature dimension.

To explicitly incorporate target information into each search token, we employ a Token Aggregator Module (TAM).
TAM first predicts normalized importance weights for the multimodal template tokens:
\begin{equation}
\boldsymbol{\eta}
=
\operatorname{Softmax}
\left(
\mathbf{T}_{f}\mathbf{W}_{t}
\right)
\in
\mathbb{R}^{N_t},
\end{equation}
where
$\mathbf{W}_{t}\in\mathbb{R}^{C\times 1}$
is a learnable projection matrix.
The compact template representation is further projected into a lightweight target embedding:
\begin{equation}
\mathbf{t}_{f}
=
\boldsymbol{\eta}^{\mathrm{T}}\mathbf{T}_{f}
\in\mathbb{R}^{C},
\qquad
\widetilde{\mathbf{t}}_{f}
=
\phi_{t}(\mathbf{t}_{f})
\in\mathbb{R}^{D_t},
\end{equation}
where $D_t$ denotes the target-embedding dimension.

The target embedding is replicated along the search-token dimension and concatenated with the projected modality features:
\begin{equation}
\mathbf{X}
=
\left[
\widetilde{\mathbf{S}}_{\mathrm{rgb}};
\widetilde{\mathbf{S}}_{\mathrm{tir}};
\operatorname{Expand}_{N_s}
\left(
\widetilde{\mathbf{t}}_{f}
\right)
\right].
\end{equation}
This construction allows every spatial token to jointly model its RGB feature, TIR feature, and target condition.

TA-HyperNet takes $\mathbf{T}_{f}$ as input and jointly generates two groups of modulation parameters:
\begin{equation}
\left\{
\boldsymbol{\gamma}_{j},
\boldsymbol{\beta}_{j}
\right\}_{j=1}^{2}
=
\mathcal{H}(\mathbf{T}_{f}),
\end{equation}
where
$\boldsymbol{\gamma}_{j},
\boldsymbol{\beta}_{j}
\in\mathbb{R}^{H}$.
Unlike dynamically generating the complete weights of the fusion network, TA-HyperNet only predicts channel-wise scaling and shifting parameters.

The two target-conditioned transformations are formulated as
\begin{equation}
\begin{aligned}
\mathbf{H}_{1}
&=
\phi_{1}(\mathbf{X})
\odot
\left(
\mathbf{1}+\boldsymbol{\gamma}_{1}
\right)
+
\boldsymbol{\beta}_{1},\\
\mathbf{H}_{2}
&=
\phi_{2}(\mathbf{H}_{1})
\odot
\left(
\mathbf{1}+\boldsymbol{\gamma}_{2}
\right)
+
\boldsymbol{\beta}_{2},
\end{aligned}
\end{equation}
where $\phi_1(\cdot)$ and $\phi_2(\cdot)$ denote linear projections, and the modulation parameters are broadcast along the token dimension.
The residual scaling form
$\mathbf{1}+\boldsymbol{\gamma}_{j}$
preserves the original feature transformation while introducing bounded target-conditioned adjustments.

The transformed features are mapped to two modality-specific logits, from which the token-wise modality weights are obtained:
\begin{equation}
\mathbf{W}
=
\operatorname{Sigmoid}
\left(
\phi_{3}(\mathbf{H}_{2})
\right)
=
\left[
\mathbf{w}_{\mathrm{rgb}},
\mathbf{w}_{\mathrm{tir}}
\right]
\in
(0,1)^{N_s\times 2}.
\end{equation}
where
$\mathbf{w}_{\mathrm{rgb}},\mathbf{w}_{\mathrm{tir}}\in(0,1)^{N_s}$.
We employ independent Sigmoid functions rather than a Softmax operation, such that the two modality weights are not forced to compete or sum to one.
This design allows both modalities to be simultaneously emphasized when they provide complementary information, or jointly suppressed when their local responses are unreliable.

Finally, the fused search representation is obtained by
\begin{equation}
\mathbf{S}_{f}
=
\mathbf{w}_{\mathrm{rgb}}\odot\mathbf{S}_{\mathrm{rgb}}
+
\mathbf{w}_{\mathrm{tir}}\odot\mathbf{S}_{\mathrm{tir}},
\end{equation}
where the modality weights are broadcast along the channel dimension.
In our implementation, $C_p=64$ and $D_t=8$.
Through target-conditioned modulation and token-wise modality weighting, TPDFM adaptively preserves target-relevant multimodal cues while suppressing unreliable local responses.

\subsection{Dynamic Spatio-temporal Calibration Module}
\label{sec:scm}

Existing methods~\cite{sttrack,cadtrack} propagate spatio-temporal tokens to exploit historical target cues.
However, recurrent propagation may also accumulate tracking noise, thereby degrading the reliability of temporal representations.
To address these issues, we propose a Dynamic Spatio-Temporal Calibration Module (DSCM), which independently calibrates and propagates the spatio-temporal tokens of each modality, as illustrated in Fig.~\ref{network}.

Let $\mathcal{M}=\{\mathrm{rgb},\mathrm{tir}\}$ denote the modality set.
For each modality $m\in\mathcal{M}$,
$\mathbf{T}_{m}^{i}\in\mathbb{R}^{N_i\times C}$,
$\mathbf{T}_{m}^{d}\in\mathbb{R}^{N_d\times C}$, and
$\mathbf{S}_{m}^{t}\in\mathbb{R}^{N_s\times C}$
denote the initial-template tokens, dynamic-template tokens, and search tokens at frame $t$, respectively.
At the first frame, the modality-specific spatio-temporal state is initialized from the corresponding initial template.
Following the global--local aggregation strategy of TRAM, a global target representation and multiple query-guided local target representations are extracted and projected into the initial spatio-temporal tokens:
\begin{equation}
\begin{array}{lcl}
(\mathbf{z}_{g,m}^{0},\mathbf{Z}_{l,m}^{0})
&=&
\operatorname{TRAM}_{p}(\mathbf{T}_{m}^{i}),\\
\mathbf{P}_{m}^{0}
&=&
\phi_{p}
\left(
[\mathbf{z}_{g,m}^{0};\mathbf{Z}_{l,m}^{0}]
\right)
\in
\mathbb{R}^{N_p\times C}.
\end{array}
\end{equation}
Here,
$\mathbf{z}_{g,m}^{0}\in\mathbb{R}^{C}$ and
$\mathbf{Z}_{l,m}^{0}\in\mathbb{R}^{(N_p-1)\times C}$
denote the global and local target representations of modality $m$, respectively, while
$\phi_p(\cdot)$ denotes a token-wise projection.
Their concatenation produces $N_p$ initial spatio-temporal tokens for each modality.

At frame $t$, TA-HyperNet takes the initial and dynamic template tokens of each modality as target conditions and generates modality-specific channel-wise scaling and shifting parameters.
The corresponding historical spatio-temporal tokens are then calibrated as
\begin{equation}
\begin{array}{lcl}
(\boldsymbol{\gamma}_{m}^{t},\boldsymbol{\beta}_{m}^{t})
&=&
\mathcal{H}_{s}
\left(
[\mathbf{T}_{m}^{i};\mathbf{T}_{m}^{d}]
\right),\\
\widehat{\mathbf{P}}_{m}^{t}
&=&
\mathbf{P}_{m}^{t-1}
\odot
\left(
\mathbf{1}+\boldsymbol{\gamma}_{m}^{t}
\right)
+
\boldsymbol{\beta}_{m}^{t}.
\end{array}
\end{equation}
where
$\boldsymbol{\gamma}_{m}^{t},\boldsymbol{\beta}_{m}^{t}\in\mathbb{R}^{C}$
and $\mathcal{H}_{s}(\cdot)$ denotes the parameter-generation head used by DSCM.
The modulation parameters are broadcast along the token dimension.
This residual modulation preserves the historical representation while adapting its channel responses to the current target appearance of the corresponding modality.

For each modality, the initial-template tokens, dynamic-template tokens, calibrated historical tokens, and current search tokens are concatenated into an independent sequence:
\begin{equation}
\mathbf{X}_{m}^{t}
=
[
\mathbf{T}_{m}^{i};
\mathbf{T}_{m}^{d};
\widehat{\mathbf{P}}_{m}^{t};
\mathbf{S}_{m}^{t}
]
\in
\mathbb{R}^{L\times C},
\end{equation}
where
$L=N_i+N_d+N_p+N_s$.
The RGB and TIR tokens are processed separately throughout DSCM, preventing the historical information of one modality from interfering with the temporal propagation of the other modality.

A self-attention layer is independently applied to each modality-specific sequence:
\begin{equation}
\begin{aligned}
\mathbf{A}_{m}^{t}
&=
\operatorname{Softmax}
\left(
\frac{
(\mathbf{X}_{m}^{t}\mathbf{W}_{Q})
(\mathbf{X}_{m}^{t}\mathbf{W}_{K})^{\mathrm{T}}
}{
\sqrt{C}
}
\right),\\
\mathbf{Y}_{m}^{t}
&=
\mathbf{X}_{m}^{t}
+
\phi_{o}
\left(
\mathbf{A}_{m}^{t}
\mathbf{X}_{m}^{t}
\mathbf{W}_{V}
\right),
\end{aligned}
\end{equation}
where
$\mathbf{W}_{Q}$,
$\mathbf{W}_{K}$, and
$\mathbf{W}_{V}$
denote the query, key, and value projections, respectively, and
$\phi_o(\cdot)$ denotes the output projection.
Finally, the updated historical tokens
$\mathbf{P}_{m}^{t}$
and propagated search features
$\widetilde{\mathbf{S}}_{m}^{t}$
are separated from
$\mathbf{Y}_{m}^{t}$.
The former is propagated to the next frame as the modality-specific temporal state, while the latter is forwarded to the subsequent multimodal fusion and prediction stages.
In this way, DSCM suppresses unreliable historical information while avoiding mutual interference between RGB and TIR temporal representations.

\begin{table*}[!t]
    \centering
    
    \setlength{\tabcolsep}{1mm}
    \renewcommand{\arraystretch}{1.0}

    \begin{tabular}{@{}c|c|c|cc|cc|cc|ccc@{}}
        \hline
        \multirow{2}{*}{Method}
        & \multirow{2}{*}{Source}
        & \multirow{2}{*}{Resolution}
        & \multicolumn{2}{c|}{GTOT}
        & \multicolumn{2}{c|}{RGBT210}
        & \multicolumn{2}{c|}{RGBT234}
        & \multicolumn{3}{c}{LasHeR}
        \\
        \cline{4-12}
        & &
        & MPR$\uparrow$ & MSR$\uparrow$
        & PR$\uparrow$ & SR$\uparrow$
        & MPR$\uparrow$ & MSR$\uparrow$
        & PR$\uparrow$ & NPR$\uparrow$ & SR$\uparrow$
        \\
        \hline

        TBSI~\cite{tbsi}
        & CVPR 2023
        & $256\times256$
        & -- & --
        & 85.3 & 62.5
        & -- & --
        & 69.2 & 65.7 & 55.6
        \\

        CKD~\cite{CKD}
        & ACM MM 2024
        & $256\times256$
        & 93.2 & 77.2
        & 88.4 & 65.2
        & 90.0 & 67.4
        & 73.2 & 69.3 & 58.1
        \\

        BAT~\cite{BAT2024}
        & AAAI 2024
        & $256\times256$
        & -- & --
        & -- & --
        & 86.8 & 64.1
        & 70.2 & -- & 56.3
        \\
        
        TATrack~\cite{TATrack}
        & AAAI 2024
        & $256\times256$
        & -- & --
        & 85.3 & 61.8
        & -- & --
        & 70.2 & 66.7 & 56.1
        \\


        US-Track~\cite{US-Track}
        & IJCAI 2024
        & $256\times256$
        & 93.4 & 78.3
        & -- & --
        & -- & --
        & -- & -- & --
        \\

        AINet~\cite{ainet}
        & AAAI 2025
        & $384\times384$
        & -- & --
        & 87.5 & 64.8
        & -- & --
        & 74.2 & 70.1 & 59.1
        \\

        STTrack~\cite{sttrack}
        & AAAI 2025
        & $256\times256$
        & -- & --
        & -- & --
        & 89.8 & 66.7
        & 76.0 & -- & 60.3
        \\

        SUTrack~\cite{SUTrack}
        & AAAI 2025
        & $384\times384$
        & -- & --
        & -- & --
        & 92.1 & 69.2
        & 75.8 & -- & 60.9
        \\

        TUMFNet~\cite{TUMFNet}
        & IJCAI 2025
        & $256\times256$
        & 95.5 & 80.2
        & 90.7 & 65.8
        & 90.8 & 67.8
        & 76.4 & 72.7 & 61.4
        \\

        XTrack~\cite{xtrack}
        & ICCV 2025
        & $256\times256$
        & -- & --
        & -- & --
        & 87.4 & 64.9
        & 69.1 & -- & 55.7
        \\



        VCT~\cite{vct2026}
        & TMM 2026
        & $384\times384$
        & 94.8 & 80.5
        & 91.4 & 64.5
        & 92.4 & 68.8
        & 77.6 & -- & 62.0
        \\

        UATrack~\cite{UATrack}
        & IJCV 2026
        & $256\times256$
        & \textbf{95.8} & 80.2
        & 92.0 & 66.6
        & 93.3 & 69.5
        & 78.5 & 74.7 & 62.6
        \\

        RAGTrack~\cite{ragtrack2026}
        & CVPR 2026
        & $256\times256$
        & -- & --
        & \textbf{93.2} & 67.1
        & 93.8 & 69.5
        & 76.8 & 73.0 & 61.1
        \\

        SCDT~\cite{SCDT}
        & CVPR 2026
        & $256\times256$
        & -- & --
        & -- & --
        & 93.1 & 69.6
        & 77.4 & -- & 61.0
        \\

        CADTrack~\cite{cadtrack}
        & AAAI 2026
        & $256\times256$
        & -- & --
        & 91.2 & 65.4
        & 92.8 & 67.7
        & 77.7 & 73.3 & 61.3
        \\

        \hline
        \textbf{PAFCNet-256}
        & Ours
        & $256\times256$
        & 95.0 & \textbf{81.1}
        & 92.4 & \textbf{67.2}
        & \textbf{94.1} & \textbf{70.1}
        & 79.8 & 75.7 & 63.4
        \\
        \textbf{PAFCNet-384}
        & Ours
        & $384\times384$
        & 95.3 & 80.2
        & 92.6 & 66.8
        & 94.0 & 69.5
        & \textbf{80.3} & \textbf{76.4} & \textbf{63.9}
        \\

        \hline
    \end{tabular}

    \caption{
    Performance on four RGBT tracking benchmarks. The best results are in \textbf{bold}.
    }
    \label{tab:overall_result}
\end{table*}

\section{Experiments}
\subsection{Implementation Details}
We adopt OSTrack~\cite{ostrack} as the baseline and initialize the model with the pretrained weights provided by DropMAE~\cite{wu2023dropmae}.
PAFCNet is implemented in PyTorch and trained on the LasHeR training set using two NVIDIA RTX 4090 GPUs. 
The model is trained for 30 epochs using
the AdamW optimizer with a learning rate of $1\times10^{-4}$ and a
weight decay of $1\times10^{-4}$.
The batch size is set to 12 and 4 for search-region resolutions of $256\times256$ and $384\times384$, respectively. 
The tracker takes an initial template, a dynamically updated template, and a single search region as inputs, and is optimized using the same training objective as OSTrack~\cite{ostrack}.
The number $N_p$ of spatio-temporal tokens is set to 64.
Further details on the training setup and other hyperparameters used by the tracker will be provided in the supplementary material.

\subsection{Comparison with State-of-the-Art Trackers}

We compare PAFCNet with recent state-of-the-art RGBT trackers on GTOT~\cite{7577747},
RGBT210~\cite{rgbt210}, RGBT234~\cite{rgbt234} and
LasHeR~\cite{lasher}. 
Following standard evaluation protocols, we report Maximum Precision Rate (MPR) and Maximum Success Rate (MSR) on GTOT and RGBT234, Precision Rate (PR) and Success Rate (SR) on RGBT210, and PR, Normalized Precision Rate (NPR), and SR on LasHeR.
The quantitative results are summarized in
Table~\ref{tab:overall_result}.

\paragraph{Evaluation on GTOT.}
On the GTOT dataset, PAFCNet-256 achieves 95.0\% MPR and 81.1\% MSR, while PAFCNet-384 obtains 95.3\% MPR and 80.2\% MSR.
Notably, PAFCNet-256 achieves the best MSR among all compared methods, surpassing the strongest competing result of 80.5\% reported by VCT by 0.6\%, while maintaining a competitive MPR of 95.0\%.
The improvement in MSR indicates that the proposed method provides more accurate target-region overlap and stable scale estimation under challenging conditions.

\paragraph{Evaluation on RGBT210.}
As shown in Table~\ref{tab:overall_result}, PAFCNet-256 achieves 92.4\% PR and 67.2\% SR.
It obtains the best SR among the compared trackers, surpassing RAGTrack by 0.1\%.
Compared with UATrack, PAFCNet-256 improves PR and SR by 0.4\% and 0.6\%, respectively.
These results demonstrate that the proposed TPDFM and DSCM enable robust target localization under diverse unseen tracking conditions.

\paragraph{Evaluation on RGBT234.}
As reported in Table~\ref{tab:overall_result}, PAFCNet-256 achieves 94.1\% MPR and 70.1\% MSR, ranking first on both metrics among all compared methods.
Compared with RAGTrack, PAFCNet-256 improves MPR and MSR by 0.3\% and 0.6\%, respectively.
Compared with UATrack, the corresponding improvements are 0.8\% in MPR and 0.6\% in MSR.
Moreover, PAFCNet-256 surpasses CADTrack by 1.3\% in MPR and 2.4\% in MSR.
The consistent improvements in both localization precision and overlap accuracy demonstrate the effectiveness of the proposed method.

\begin{figure}[!htb]
\centering
\includegraphics[width=3.1in]{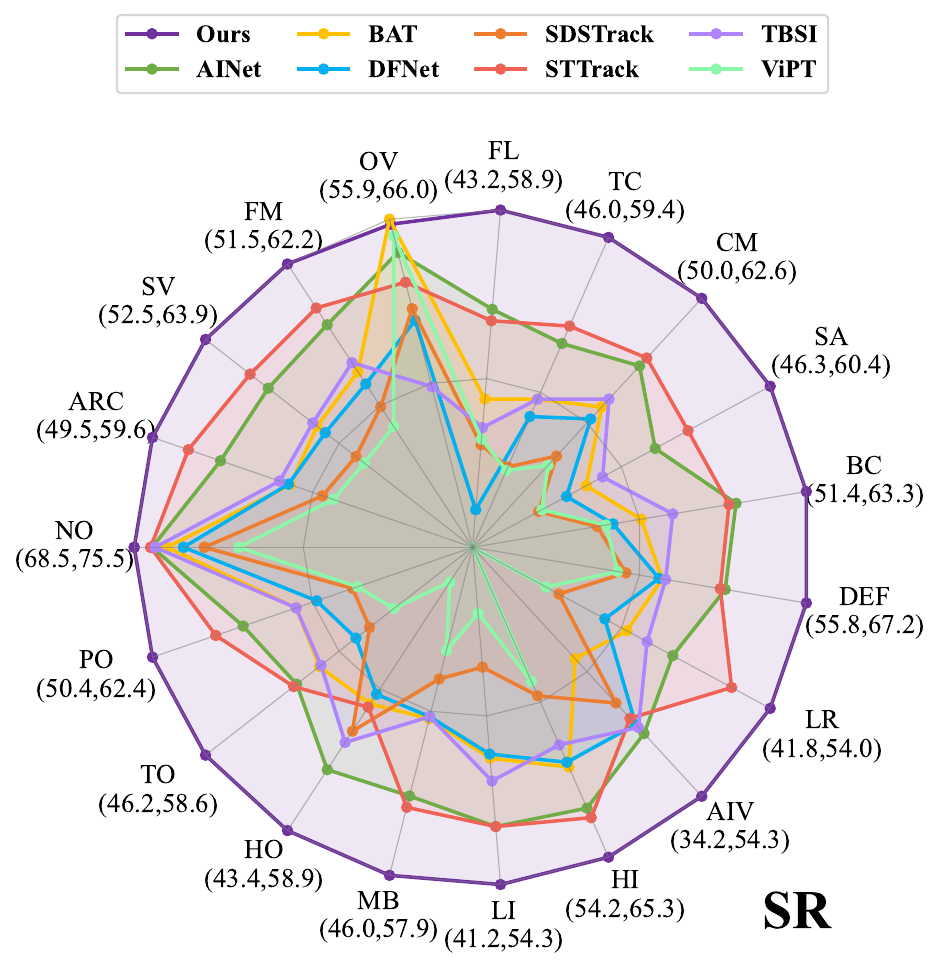}
\caption{Attribute-based evaluation on the LasHeR dataset.}
\label{SR}
\end{figure}

\paragraph{Evaluation on LasHeR.}
%
As shown in Table~\ref{tab:overall_result}, PAFCNet-384 achieves 80.3\% PR, 76.4\% NPR, and 63.9\% SR, ranking first across all three metrics among the compared methods.
Compared with UATrack, the strongest competing tracker with complete results, PAFCNet-384 improves PR, NPR, and SR by 1.8\%, 1.7\%, and 1.3\%, respectively.
It also surpasses CADTrack by 2.6\% in PR, 3.1\% in NPR, and 2.6\% in SR.
%
%
We further present the attribute-based evaluation results on the LasHeR dataset, as shown in Fig.~\ref{SR}.
%
%
TPDFM effectively exploits complementary multimodal information, yielding notable gains under low illumination (LI), high illumination (HI), and abrupt illumination variation (AIV).
Meanwhile, DSCM improves the reliability of propagated spatio-temporal information, thereby enhancing robustness against similar appearance (SA), partial occlusion (PO), total occlusion (TO), and motion blur (MB).

\begin{table}[!ht]
\centering
\setlength{\tabcolsep}{1.2mm}
\renewcommand{\arraystretch}{1.0}
\begin{tabular}{@{}lccccc@{}}
\toprule
\multirow{2}{*}{Method}
& \multicolumn{2}{c}{RGBT234}
& \multicolumn{3}{c}{LasHeR} \\
\cmidrule(lr){2-3}
\cmidrule(lr){4-6}
& MPR$\uparrow$ & MSR$\uparrow$
& PR$\uparrow$ & NPR$\uparrow$ & SR$\uparrow$ \\
\midrule
Baseline
& 89.4 & 66.5
& 71.9 & 68.3 & 57.8 \\

+ Template update
& 90.5 & 67.5
& 75.4 & 71.5 & 59.9 \\

+ DSCM
& 92.1 & 69.2
& 77.9 & 73.8 & 61.6 \\

+ TPDFM (Full Model)
& \textbf{94.1} & \textbf{70.1}
& \textbf{79.8} & \textbf{75.7} & \textbf{63.4} \\

\bottomrule
\end{tabular}
\caption{Component ablation studies on RGBT234 and LasHeR.}
\label{tab:ablation_components}
\end{table}

\subsection{Ablation Studies}
\label{sec:ablation}

\paragraph{Component Analysis.}
As shown in Table~\ref{tab:ablation_components}, progressively introducing template updating, DSCM, and TPDFM consistently improves tracking performance on both benchmarks.
Compared with the template-update variant, DSCM improves MPR and MSR on RGBT234 by 1.6\% and 1.7\%, respectively, while increasing PR, NPR, and SR on LasHeR by 2.5\%, 2.3\%, and 1.7\%.
Further incorporating TPDFM yields additional gains of 2.0\% MPR and 0.9\% MSR on RGBT234, together with improvements of 1.9\%, 1.9\%, and 1.8\% in PR, NPR, and SR on LasHeR.
Overall, the full model outperforms the baseline by 4.7\% MPR and 3.6\% MSR on RGBT234, and by 7.9\% PR, 7.4\% NPR, and 5.6\% SR on LasHeR, demonstrating the effectiveness and complementarity of DSCM and TPDFM.

\begin{table}[!ht]
\centering
\setlength{\tabcolsep}{2pt}
\renewcommand{\arraystretch}{1.0}
\begin{tabular}{@{}lcccccc@{}}
\toprule
Paradigm
& PR$\uparrow$
& NPR$\uparrow$
& SR$\uparrow$
& Params$\downarrow$
& FLOPs$\downarrow$
& FPS$\uparrow$ \\
\midrule

SF
& 77.6
& 73.4
& 61.7
& \textbf{106.3M}
& \textbf{70.9G}
& \textbf{50} \\

DAF
& 78.4
& 74.5
& 62.5
& 108.7M
& 71.4G
& 45 \\

TPDFM
& \textbf{79.8}
& \textbf{75.7}
& \textbf{63.4}
& 111.9M
& 70.9G
& 49 \\

\bottomrule
\end{tabular}
\caption{
Comparison of different fusion paradigms on LasHeR.
}
\label{tab:fusion_paradigms}
\end{table}

\paragraph{Comparison of Multimodal Fusion Paradigms.}
We compare the proposed TPDFM with static fusion (SF) and dynamic-architecture fusion (DAF), as shown in Table~\ref{tab:fusion_paradigms}.
SF replaces the target-conditioned parameters generated by TA-HyperNet with fixed learnable parameters.
DAF adopts a router-controlled mechanism similar to AFTER~\cite{AFTER}, which dynamically selects and combines predefined fusion paths.
TPDFM achieves the best performance with 79.8\% PR, 75.7\% NPR, and 63.4\% SR, outperforming SF by 2.2\%, 2.3\%, and 1.7\%, and DAF by 1.4\%, 1.2\%, and 0.9\%, respectively.
These results demonstrate the advantage of target-conditioned dynamic parameterization over fixed-parameter fusion and dynamic path selection.
More experimental details are provided in the supplementary material.

\begin{table}[htbp]
    \centering
    \renewcommand{\arraystretch}{0.8}
    \setlength{\tabcolsep}{3.2pt}
    \begin{tabular}{cccccccc}
        \toprule
        \multicolumn{3}{c}{Inserting Layers}
        & \multicolumn{2}{c}{RGBT234}
        & \multicolumn{3}{c}{LasHeR} \\
        \cmidrule(lr){1-3}
        \cmidrule(lr){4-5}
        \cmidrule(lr){6-8}
        10 & 11 & 12
        & MPR$\uparrow$ & MSR$\uparrow$
        & PR$\uparrow$ & NPR$\uparrow$ & SR$\uparrow$ \\
        \midrule
        & & &
        90.5 & 67.5
        & 75.4 & 71.5 & 59.9 \\

        \checkmark & & &
        92.8 & 69.4
        & 77.3 & 73.5 & 61.7 \\

        \checkmark & \checkmark & &
        93.1 & 69.6
        & 79.0 & 75.1 & 62.9 \\

        \checkmark & \checkmark & \checkmark
        & \textbf{94.1} & \textbf{70.1}
        & \textbf{79.8} & \textbf{75.7} & \textbf{63.4} \\
        \bottomrule
    \end{tabular}
    \caption{Effect of inserting TPDFM into different Transformer layers on RGBT234 and LasHeR.}
    \label{tab:inserting_layers}
\end{table}

\paragraph{Effect of Insertion Layers.}
We investigate the effect of inserting TPDFM into different Transformer layers on RGBT234 and LasHeR test set.
As shown in Table~\ref{tab:inserting_layers}, progressively inserting TPDFM into the final three Transformer layers consistently improves tracking performance.
Using TPDFM only in the 10th layer increases PR, NPR, and SR by 1.9\%, 2.0\%, and 1.8\%, respectively.
Extending it to the 10th and 11th layers further improves the results to 79.0\% PR, 75.1\% NPR, and 62.9\% SR.
The best performance is achieved by inserting TPDFM into all three layers, yielding 79.8\% PR, 75.7\% NPR, and 63.4\% SR.

\begin{table}[!ht]
\centering

\setlength{\tabcolsep}{3mm}
\renewcommand{\arraystretch}{1.0}
\begin{tabular}{@{}cccc@{}}
\toprule
Number of Tokens & PR$\uparrow$ & NPR$\uparrow$ & SR$\uparrow$ \\
\midrule
16  & 78.9 & 74.8 & 62.8 \\
32  & 79.3 & 75.4 & 63.1 \\
64  & \textbf{79.8} & \textbf{75.7} & \textbf{63.4} \\
128 & 78.6 & 74.7 & 62.6 \\
\bottomrule
\end{tabular}
\caption{Effect of the Number of Spatio-Temporal Tokens.}
\label{tab:token_number}
\end{table}

\paragraph{Effect of the Number of Spatio-Temporal Tokens.}
We investigate the effect of the number of spatio-temporal tokens on the LasHeR test set.
As shown in Table~\ref{tab:token_number}, increasing the token number from 16 to 64 consistently improves tracking performance, with 64 tokens achieving the best results of 79.8\% PR, 75.7\% NPR, and 63.4\% SR.
However, further increasing the number to 128 degrades all three metrics, suggesting that excessive tokens may introduce redundant temporal information and increase modeling difficulty.
Therefore, we adopt 64 spatio-temporal tokens as the default setting, which provides the best balance between representation capability and tracking performance.

\begin{figure}[!htb]
\centering
\includegraphics[width=3.1in]{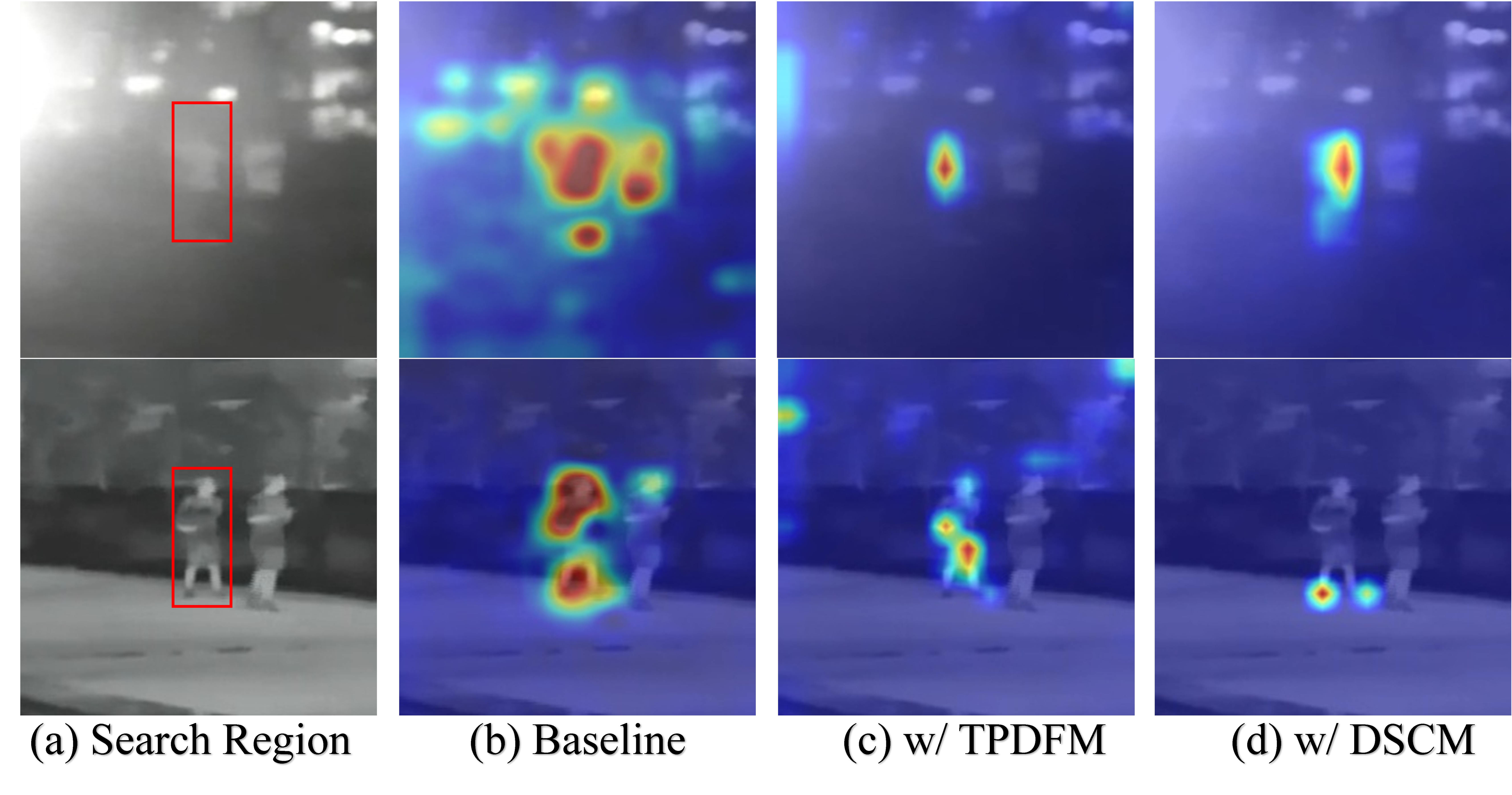}
\caption{Attention evolution for RGB (top) and TIR (bottom).}
\label{attn}
\end{figure}

\paragraph{Visualization Analysis.}
To qualitatively analyze the effects of TPDFM and DSCM, we visualize the attention maps of the RGB and TIR branches in Fig.~\ref{attn}.
The baseline produces dispersed responses and is easily affected by background interference.
With TPDFM, the attention responses become more concentrated on the target, demonstrating that target-conditioned fusion effectively enhances target-relevant multimodal information.
DSCM further suppresses distracting responses and produces more compact target activation by calibrating the propagated spatio-temporal tokens.

\section{Conclusion}
In this work, we propose PAFCNet, a Parameter-Dynamic Adaptive Fusion and Calibration Network for RGBT tracking.
Unlike conventional methods with fixed fusion functions, we first develop a Target-adaptive Hypernetwork (TA-HyperNet) to generate target-conditioned parameters from template representations.
Building on TA-HyperNet, the proposed target-aware parameter-dynamic fusion module adaptively adjusts multimodal fusion according to variations in target appearance and scene conditions.
We further introduce a dynamic spatio-temporal calibration module to calibrate propagated spatio-temporal tokens and suppress unreliable historical information.
Extensive experiments on multiple benchmarks demonstrate the effectiveness of the proposed framework.

\bibliography{aaai2027}


\end{document}